%% file: main_submission.tex
\documentclass[runningheads]{llncs}
\usepackage{graphicx}
\graphicspath{{../}}  % hangg: bypass local compiling issues.
\usepackage{comment}
\usepackage{times}
\usepackage{epsfig}
\usepackage{graphicx}
\usepackage{amsmath}
\usepackage{amssymb}
\usepackage{bm}
\usepackage{booktabs}
\usepackage{gensymb}
\usepackage{wrapfig}
\usepackage{mathtools}
\usepackage{color}
\usepackage{xcolor}
\usepackage{subcaption}
\usepackage[width=122mm,left=12mm,paperwidth=146mm,height=193mm,top=12mm,paperheight=217mm]{geometry}

\usepackage{verbatim}
\usepackage[
    pagebackref=true,
    breaklinks=true,
    letterpaper=true,
    colorlinks,
    bookmarks=false
]{hyperref}

\newcommand{\figref}[1]{Figure~\ref{#1}}
\newcommand{\figlabel}[1]{\label{#1}}

\newcommand{\boldstart}[1]{\noindent\textbf{#1}}

\begin{document}
\pagestyle{headings}
\mainmatter
\def\ECCVSubNumber{1448}

\title{Long-term Human Motion Prediction\\ with Scene Context}

\begin{comment}
\titlerunning{ECCV-20 submission ID \ECCVSubNumber}
\authorrunning{ECCV-20 submission ID \ECCVSubNumber}
\author{Anonymous ECCV submission}
\institute{Paper ID \ECCVSubNumber}
\end{comment}
%******************

\titlerunning{Long-term Human Motion Prediction with Scene Context}
\author{Zhe Cao\inst{1}%\thanks{Part of this work is done when ZC interned at Facebook Reality Labs, Sausalito, USA.} 
\and
Hang Gao\inst{1} \and
Karttikeya Mangalam\inst{1} \and 
Qi-Zhi Cai\inst{2} \and \\
Minh Vo%\inst{3} 
\and 
Jitendra Malik\inst{1}
}
\institute{UC Berkeley \and
Nanjing University %\and
%Facebook Reality Labs \\
}
\authorrunning{Cao et al.}

\maketitle

\begin{abstract}
Human movement is goal-directed and influenced by the spatial layout of the objects in the scene. To plan future human motion, it is crucial to perceive the environment -- imagine how hard it is to navigate a new room with lights off.
Existing works on predicting human motion do not pay attention to the scene context and thus struggle in long-term prediction.
In this work, we propose a novel three-stage framework that exploits scene context to tackle this task.
Given a single scene image and 2D pose histories, our method first samples multiple human motion goals, then plans 3D human paths towards each goal, and finally predicts 3D human pose sequences following each path.
For stable training and rigorous evaluation, we contribute a diverse synthetic dataset with clean annotations. In both synthetic and real datasets, our method shows consistent quantitative and qualitative improvements over existing methods. Project page: {\textit{ \small \url{https://people.eecs.berkeley.edu/~zhecao/hmp/index.html}}}

\end{abstract}

\begin{center}
    \newcommand{\teaserwidth}{1.0\textwidth}
    \centerline{
        \includegraphics[width=\teaserwidth]{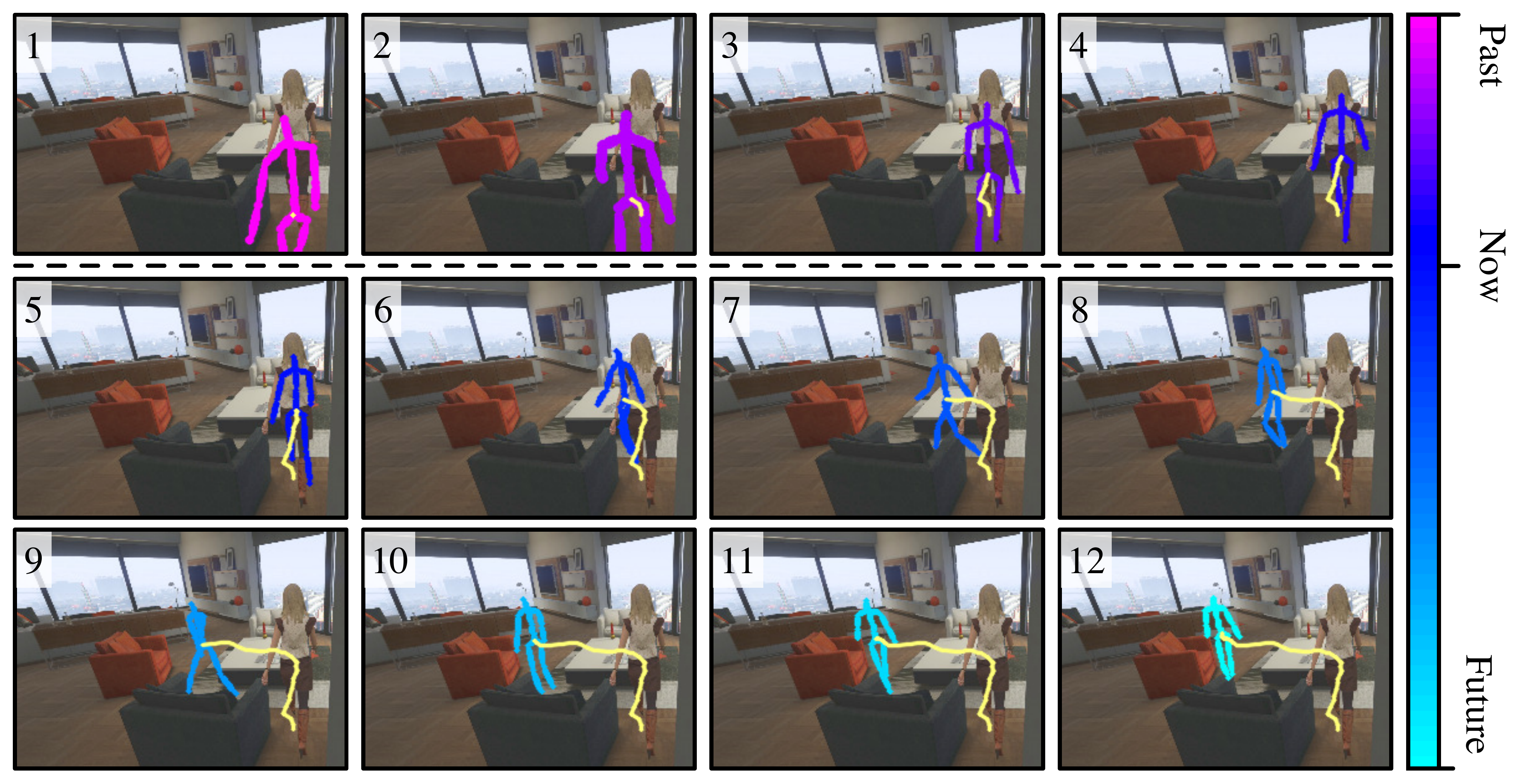}
     }
    \captionof{figure}{\small \textbf{
        Long-term 3D human motion prediction}.
        Given a single scene image and 2D pose histories (the 1st row), we aim to predict long-term 3D human motion (projected on the image, shown in the 2-3rd rows) influenced by scene. The human path is visualized as a yellow line.
    }
    \figlabel{pipeline}
\end{center}

\input{hmp_0intro.tex}
\input{hmp_1related.tex}
\input{hmp_2approach.tex}\label{sec:approach}
\input{hmp_3dataset.tex}
\input{hmp_4experiments.tex}

\input{hmp_5conclusions.tex}
\newpage

\begin{figure*}[!ht]
    \centering
    \includegraphics[width=1.0\textwidth]{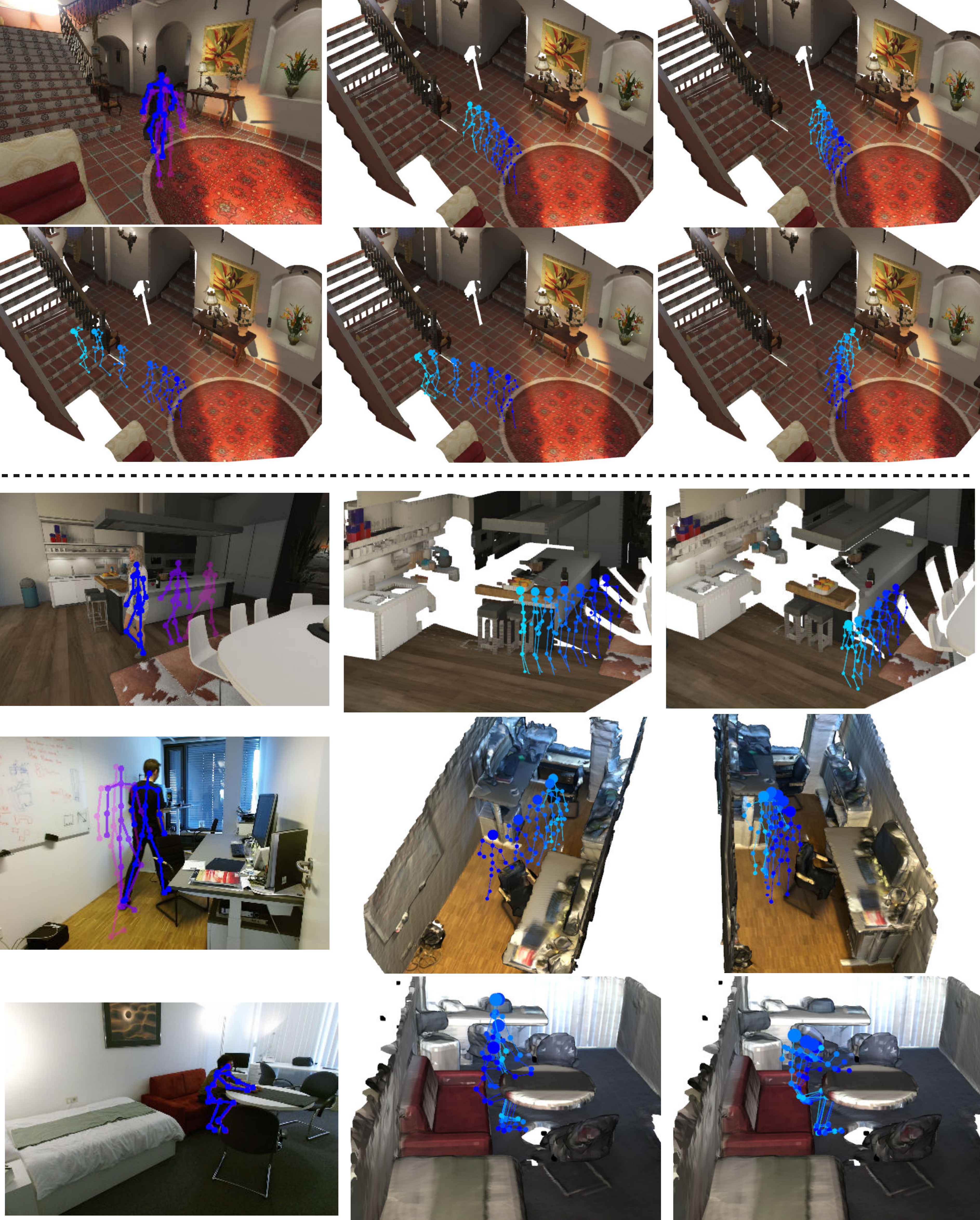} \\
    \caption{\small \textbf{Qualitative results on long-term stochastic prediction}. In each example, we first show the input image with 2D pose histories and then our stochastic predictions. In the first example (1st and 2nd row), we show five different future human movement predictions by sampling different human ``goals''. Depending on his intention, the person can choose to turn left to climb upstairs; he may also go straight through the hallway or turn right to fetch some items off the table. For the following examples at each row, we only show two stochastic predictions per example due to space limitation.
    Our method can generate diverse human motion, e.g., turning left/right, walking straight, taking a u-turn, standing up from sitting, and laying back on the sofa.
    }
    \vspace{-5pt}
    \figlabel{qualitative}
\end{figure*}

\bibliographystyle{splncs04}
\bibliography{hmp_references}

\newpage
\appendix
\input{main_supplement}

\end{document}

%% file: hmp_0intro.tex
\section{Introduction}
Figure~\ref{pipeline} shows the image of a typical indoor scene. Overlaid on this image is the pose trajectory of a person, depicted here by renderings of her body skeleton over time instants, where Frames 1-3 are in the past, Frame 4 is the present, and Frames 5-12 are in the future.
In this paper, we study the following problem: \emph{Given the scene image and the person’s past pose and location history in 2D, predict her future poses and locations.}

Human movement is goal-directed and influenced by the spatial layout of the objects in the scene. For example, the person may be heading towards the window, and will find a path through the space avoiding collisions with various objects that might be in the way. Or perhaps a person approaches a chair with the intention to sit on it, and will adopt an appropriate path and pose sequence to achieve such a goal efficiently. We seek to understand such goal-directed, spatially contextualized human behavior, which we have formalized as a pose sequence and location prediction task.

With the advent of deep learning, there has been remarkable progress on the task of predicting human pose sequences ~\cite{fragkiadaki2015recurrent,martinez2017human,wei2019motion,zhang2019phd}. However, these frameworks do not pay attention to scene context. As a representative example, Zhang et al.~\cite{zhang2019phd} detect the human bounding boxes across multiple time instances and derive their predictive signal from the evolving appearance of the human figure, but do not make use of the background image. Given this limitation, the predictions tend to be short-term~(around 1 second), and local in space, e.g., walking in the same spot without global movement. If we want to make predictions that encompass bigger spatiotemporal neighborhoods, we need to make predictions conditioned on the scene context.

We make the following philosophical choices:
(1) To understand long term behavior, we must reason in terms of goals. In the setting of moving through space, the goals could be represented by the destination points in the image. We allow multi-modality by generating multiple hypotheses of human movement ``goals'', represented by 2D destinations in the image space.
(2) Instead of taking a 3D path planning approach as in the classical robotics literature~\cite{alexopoulos1992path,lavalle2006planning}, we approach the construction of likely human motions as a learning problem by constructing a convolutional network to implicitly learn the scene constraints from lots of human-scene interaction videos. We represent the scene using 2D images. 

Specifically, we propose a learning framework that factorizes this task into three sequential stages as shown in Figure~\ref{fig2}. Our model sequentially predicts the motion goals, plans the 3D paths following each goal and finally generates the 3D poses.
In Section~\ref{sec:exp}, we demonstrate our model not only outperforms existing methods quantitatively but also generates more visually plausible 3D future motion.

To train such a learning system, we contribute a large-scale synthetic dataset focusing on human-scene interaction.
Existing real datasets on 3D human motion have either contrived environment~\cite{ionescu2013human3,wang2019geometric}, relatively noisy 3D annotations~\cite{savva2016pigraphs}, or limited motion range due to the depth sensor~\cite{hassan2019resolving,savva2016pigraphs}.
This motivates us to collect a diverse synthetic dataset with clean 3D annotations. We turn the Grand Theft Auto (GTA) gaming engine into an automatic data pipeline with control over different actors, scenes, cameras, lighting conditions, and motions.
We collect over one million HD resolution RGB-D frames with 3D annotations which we discuss in detail in Section~\ref{GTA}.
Pre-training on our dataset stabilizes training and improves prediction performance on real dataset~\cite{hassan2019resolving}.

In summary, our key contributions are the following:
(1) We formulate a new task of long-term 3D human motion prediction with scene context in terms of 3D poses and 3D locations.
(2) We develop a novel three-stage computational framework that utilizes scene context for goal-oriented motion prediction, which outperforms existing methods both quantitatively and qualitatively.
(3) We contribute a new synthetic dataset with diverse recordings of human-scene interaction and clean annotations. 

\begin{figure}[t]
    \centering
    \includegraphics[width=1.0\textwidth]{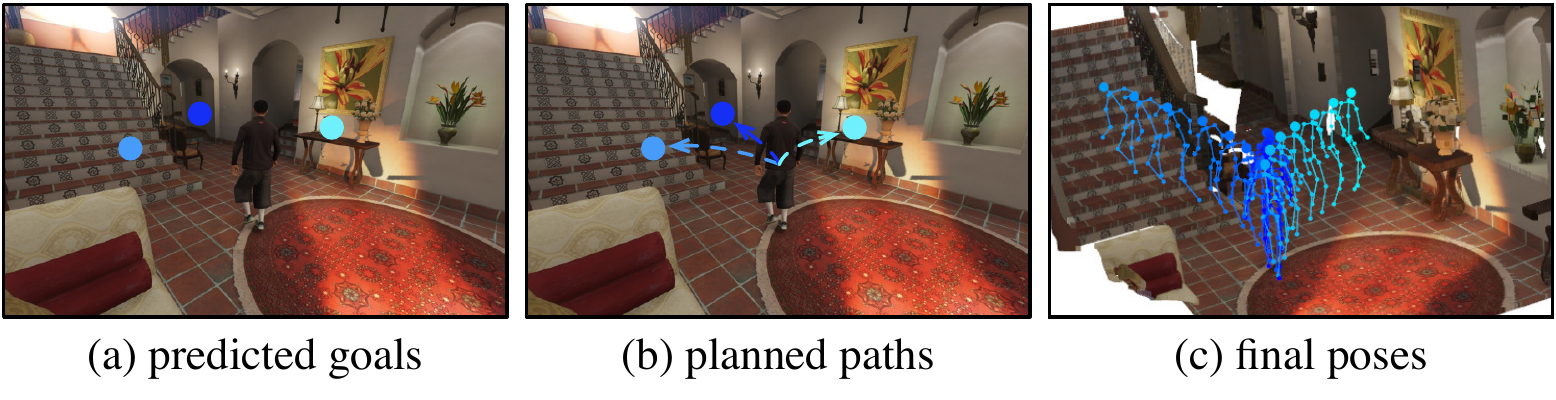}
    \vspace{-1.5em}
    \caption{\small{\textbf{Overall pipeline}.
    Given a single scene image and 2D pose histories, our method first samples (a) multiple possible future 2D destinations. We then predict the (b) 3D human path towards each destination. Finally, our model generates (c) 3D human pose sequences following paths, visualized with the ground-truth scene point cloud.}}
    \figlabel{fig2}
    \vspace{-5pt}
\end{figure}

%% file: hmp_1related.tex
\section{Related Work}
Predicting future human motion under real-world social context and scene constraints is a long-standing problem~\cite{alahi2016social,gupta2018social,helbing1995social,kitani2012activity,sadeghian2019sophie}. Due to its complexity, most of the current approaches can be classified into global trajectory prediction and local pose prediction. We connect these two components in a single framework for long-term scene-aware future human motion prediction.

\vspace{2pt}
\boldstart{Global trajectory prediction:} Early approaches in trajectory prediction model the effect of social-scene interactions using physical forces~\cite{helbing1995social}, continuum dynamics~\cite{treuille2006continuum},  Hidden Markov model~\cite{kitani2012activity}, or game theory~\cite{ma2017forecasting}. Many of these approaches achieve competitive
results even on modern pedestrian datasets~\cite{lerner2007crowds,pellegrini2009you}. With the resurgence of neural nets, data-driven prediction paradigm that captures multi-modal interaction between the scene and its agents becomes more dominant~\cite{alahi2016social,alahi2014socially,chai2019multipath,gupta2018social,makansi2019overcoming,sadeghian2019sophie,tai2018socially,yu2018one}. 
Similar to our method, they model the influence of the scene implicitly. 
However, unlike our formulation that considers images from diverse camera viewpoints, they make the key assumption of the bird-eye view image or known 3D information~\cite{alahi2016social,gupta2018social,kitani2012activity,sadeghian2019sophie}. 

\vspace{2pt}
\boldstart{Local pose prediction:} 
Similar to trajectory prediction, there has been plenty of interest in predicting future pose from image sequences both in the form of image generation~\cite{villegas2017learning,zhao2018learning}, 2D pose ~\cite{chao2017forecasting,walker2017pose}, and 3D pose~\cite{chiu2019action,ghosh2017learning,weng2019photo,zhang2019phd}. 
These methods exploit the local image context around the human to guide the future pose generation but do not pay attention to the background image or the global scene context. 
Approaches that focus on predicting 3D pose from 3D pose history also exist and are heavily used in motion tracking~\cite{elhayek2012spatio,vo2016spatiotemporal}. The goal is to learn 3D pose prior conditioning on the past motion using techniques such as Graphical Models~\cite{brand2000style}, linear dynamical systems~\cite{pavlovic2001learning}, trajectory basis~\cite{akhter2009nonrigid,akhter2012bilinear}, or Gaussian Process latent variable models~ \cite{tay2008modelling,urtasun2008topologically,wang2007gaussian,wang2007multifactor}, and more recently neural networks such as recurrent nets~\cite{fragkiadaki2015recurrent,jain2016structural,li2018auto,martinez2017human,pavllo2018quaternet}, temporal convolution nets~\cite{hernandez2019human,holden2015learning,Li_2018_CVPR}, or graph convolution net in frequency space~\cite{wei2019motion}. However, since these methods completely ignore the image context, the predicted human motion may not be consistent with the scene, i.e, waling through the wall. In contrast, we propose to utilize the scene context for future human motion prediction. This is similar in spirit to iMapper~\cite{monszpart2018imapper}. However, this approach relies on computationally expensive offline optimization to jointly reason about the scene and the human motion. Currently, there is no learning-based method that holistically models the scene context and human pose for more than a single time instance~\cite{chen2019holistic++,lee2018context,li2019putting,wang2017binge,wang2020predicting}.

\vspace{2pt}
\boldstart{3D Human Motion Dataset} Training high capacity neural models requires large-scale and diverse training data. Existing human motion capture datasets either contain no environment~\cite{CMUMocap}, contrive environment~\cite{ionescu2013human3,wang2019geometric}, or in the outdoor setting without 3D annotation~\cite{vonMarcard2018}. Human motion datasets with 3D scenes are often much smaller and have relatively noisy 3D human poses~\cite{hassan2019resolving,savva2016pigraphs} due to the limitations of the depth sensor. To circumvent such problems, researchers exploit the interface between the game engine and the graphics rendering system to collect large-scale synthetic datasets~\cite{fabbri2018learning,krahenbuhl2018free}. Our effort on synthetic training data generation is a consolidation of such work to the new task of future human motion prediction with scene context.

%% file: hmp_2approach.tex
\section{Approach}
In this work, we focus on long-term 3D human motion prediction that is goal-directed and is under the influence of scene context. We approach this problem by constructing a learning framework that factorizes long-term human motions into modeling their potential goal, planing 3D path and pose sequence, as shown in Figure~\ref{network}.
Concretely, given a $N$-step 2D human pose history~${\bf X}_{1:N}$ and an 2D image\footnote[1]{We choose to represent the scene by RGB images rather than RGBD scans because they are more readily available in many practical applications.} of the scene~${\bf I}$ (the $N$th video frame in our case), we want to predict the next~$T$-step 3D human poses together with their locations, denoted by a sequence ${\bf Y}_{N+1:N+T}$.
We assume a known human skeleton consists of $J$ keypoints, such that ${\bf X} \in \mathbb{R}^{J\times2}, {\bf Y} \in \mathbb{R}^{J\times3}$.
We also assume a known camera model parameterized by its intrinsic matrix ${\bf K} \in \mathbb{R}^3$.
To denote a specific keypoint position, we use the superscript of its index in the skeleton, e.g., ${\bf X}^r$ refers to the 2D location of the human center (torso) indexed by $r \in [1, J]$.

We motivate and elaborate our modular design for each stage in the rest of the section.
Specifically, \textit{GoalNet} learns to predict multiple possible human motion goals, represented as 2D destinations in the image space, based on a 2D pose history and the scene image.
Next, \textit{PathNet} learns to plan a 3D path towards each goal -- the 3D location sequence of the human center (torso) -- in conjunction with the scene context.
Finally, \textit{PoseNet} predicts 3D human poses at each time step following the predicted 3D path.
In this way, the resulting 3D human motion has global movement and is more plausible considering the surrounding scene. 

Thanks to this modular design, our model can have either deterministic or stochastic predictions. When deploying GoalNet, our model can sample multiple destinations, which results in stochastic prediction of future human motion. If not deploying GoalNet, our model generates single-mode prediction instead. We discuss them in more detail in the rest of the section and evaluate both predictions in our experiments.

\begin{figure*}[t]
    \centering
    \includegraphics[width=1.0\textwidth]{figures/sec3/network_v5.pdf} \\
    \caption{\small \textbf{Network architecture}. Our pipeline contains three stages: \textit{GoalNet} predicts 2D motion destinations of the human based on the reference image and 2D pose heatmaps (Section~\ref{sec:GoalNet}); \textit{PathNet} plans the 3D global path of the human with the input of 2D heatmaps, 2D destination, and the image (Section~\ref{sec:pathnet}); \textit{PoseNet} predicts 3D global human motion, i.e., the 3D human pose sequences, following the predicted path (Section~\ref{sec:posenet}).
    }
    \figlabel{network}
    \vspace{-5pt}
\end{figure*}

\subsection{\textit{GoalNet}: Predicting 2D Path Destination} \label{sec:GoalNet}
To understand long-term human motion, we must reason in terms of goals.
Instead of employing autoregressive models to generate poses step-by-step, we seek to first directly predict the destination of the motion in the image space.
We allow our model to express uncertainty of human motion by learning a distribution of possible motion destinations, instead of a single hypothesis.
This gives rise to our GoalNet denoted as~$\mathcal{F}$ for sampling plausible 2D path destination.

GoalNet learns a distribution of possible 2D destinations $\{\hat{{\bf X}}^{r}_{N+T}\}$ at the end of the time horizon conditioned on the 2D pose history ${\bf X}_{1:N}$ and the scene image ${\bf I}$. We parametrize each human keypoint ${\bf X}^j$ by a heatmap channel ${\bf H}^j$ which preserves spatial correlation with the image context.

We employ GoalNet as a conditional variational auto-encoder~\cite{kingma2013auto}. The model encodes the inputs into a latent ${\bf z}$-space, from which we sample a random ${\bf z}$ vector for decoding and predicting the target destination positions.
Formally, we have
\begin{equation}
    {\bf z} \sim \mathcal{Q}({\bf z}|{\bf H}_{1:N}^{1:J},{\bf I})
    \equiv
    \mathcal{N}(\bm{\mu}, \bm{\sigma})
    \text{, where }\bm{\mu}, \bm{\sigma} = \mathcal{F}_{\text{enc}} ({\bf H}_{1:N}^{1:J}, {\bf I}).
\end{equation}
In this way, we estimate a variational posterior $\mathcal{Q}$ by assuming a Gaussian information bottleneck using the decoder.
Next, given a sampled ${\bf z}$ latent vector, we learn to predict our target destination heatmap with our GoalNet decoder,
\begin{equation}
    \hat{{\bf H}}^r_{N+T} = \mathcal{F}_{\text{dec}} ({\bf z}, {\bf I}),
\end{equation}
where we additionally condition the decoding process on the scene image. We use soft-argmax~\cite{sun2018integral} to extract the 2D human motion destination $\hat{{\bf X}}^r_{N+T}$ from this heatmap $\hat{{\bf H}}^r_{N+T}$.
We choose to use soft-argmax operation because it is differentiable and can produce sub-pixel locations.
By constructing GoalNet, we have
\begin{equation}
    \hat{{\bf H}}^r_{N+T}
    =
    \mathcal{F}({\bf I}, {\bf H}_{1:N}^{1:J}).
\end{equation}
We train GoalNet by minimizing two objectives: (1) the destination prediction error and (2) the KL-divergence between the estimated variational posterior $\mathcal{Q}$ and a normal distribution $\mathcal{N}(\bm{0},\bm{1})$:
\begin{eqnarray}
    \begin{aligned}
    L_{\text{dest2D}}
    &=
    \|{\bf X}^{r}_{N+T}
    -
    \hat{{\bf X}}^{r}_{N+T}\|_1,
    \\
    L_{\text{KL}}
    &=
    \text{KL}
    \big[
        \mathcal{Q}
        \big(
        {\bf z}
        |
        {\bf H}_{1:N}^{1:J}
        ,
        {\bf I}
        \big)
        ||
        \mathcal{N}(0, 1)
    \big],
    \end{aligned}
\end{eqnarray}
where we weigh equally between them.
During testing, our GoalNet is able to sample a set of latent variables $\{{\bf z}\}$ from $\mathcal{N}(\bm{0}, \bm{1})$ and map them to multiple plausible 2D destinations $\{\hat{{\bf H}}^r_{N+T}\}$.

\subsection{\textit{PathNet}: Planning 3D Path towards Destination} \label{sec:pathnet}
With predicted destinations in the image space, our method further predicts 3D paths (human center locations per timestep) towards each destination.
The destination determines where to move while the scene context determines how to move.
We design a network that exploits both the 2D destination and the image for future 3D path planning.
A key design choice we make here is that, instead of directly regressing 3D global coordinate values of human paths, we represent the 3D path as a combination of 2D path heatmaps and the depth values of the human center over time.
This 3D path representation facilitates training as validated in our experiments (Section~\ref{quantitative}).

As shown in \figref{network}, our PathNet $\Phi$ takes the scene image ${\bf I}$, the 2D pose history ${\bf H}_{1:N}^{1:J}$, and the 2D destination assignment $\hat{\bf H}_{N + T}^r$ as inputs, and predicts global 3D path represented as $(\hat{\bf H}_{N+1:N+T}^r, \hat{\bf d}_{1:N+T}^r)$, where $\hat{d}_t^r\in \mathbb{R}$ denotes the depth of human center at time $t$:
\begin{equation} \label{eq:pathnet}
    \hat{\bf H}_{N+1:N+T}^r, \hat{\bf d}_{1:N+T}^r
    =
    \Phi({\bf I}, {\bf X}_{1:N}^{1:J}, {\bf X}_{N + T}^r).
\end{equation}
We use soft-argmax to extract the resulting 2D path $\hat{\bf X}_{N+1:N+T}^r$ from predicted heatmaps $\hat{\bf H}_{N+1:N+T}^r$.
Finally, we obtain the 3D path $\hat{{\bf Y}}_{1:N+T}^r$ by back-projecting the 2D path into the 3D camera coordinate frame using the human center depth $\hat{\bf d}_{1:N+T}^r$ and camera intrinsics ${\bf K}$.

We use Hourglass54~\cite{law2019cornernet,newell2016stacked} as the backbone of PathNet to encode both the input image and 2D pose heatmaps. The network has two branches where the first branch predicts 2D path heatmaps and the second branch predicts the depth of the human torso.
We describe the network architecture with more details in Appendix~\ref{sec:supp-network}.

We train our PathNet using two supervisions. We supervise our path predictions with ground-truth 2D heatmaps:
\begin{equation}
    L_{\text{path2D}}
    =
    \|{\bf X}^{r}_{N+1:N+T}
        -
        \hat{\bf X}^{r}_{N+1:N+T} \|_1.
\end{equation}
We also supervise path predictions with 3D path coordinates, while encouraging smooth predictions by penalizing large positional changes between consecutive frames:
\begin{equation}
    L_{\text{path3D}}
    =
    \| {\bf Y}^{r}_{1:N+T}
        -
        \hat{\bf Y}^{r}_{1:N+T} \|_1
    +
    \| \hat{\bf Y}^{r}_{1:N+T-1} 
        -
        \hat{\bf Y}^{r}_{2:N+T} \|_1
    .
\end{equation}
These losses are summed together with equal weight as the final training loss. During training, we use the ground-truth destination to train our PathNet, while during testing, we can use predictions from the GoalNet.

The GoalNet and PathNet we describe so far enable sampling multiple 3D paths during inference.
We thus refer to it as the stochastic mode of the model.
The modular design of GoalNet and PathNet is flexible.
By removing GoalNet and input ${\bf X}_{N + T}^r$ from Equation~\ref{eq:pathnet}, we can directly use PathNet to produce deterministic 3D path predictions.
We study these two modes, deterministic and stochastic mode, in our experiments.

\subsection{\textit{PoseNet}: Generating 3D Pose following Path}
\label{sec:posenet}
With the predicted 3D path~$\hat{{\bf Y}}^{r}_{1:N+T}$ and 2D pose history ${\bf X}_{1:N}$, we use the transformer network~\cite{vaswani2017attention} as our PoseNet~$\Psi$ to predict 3D poses following such path. %In this final stage, the model learns to lift 2D pose history into 3D and predict 3D future human movement.
Instead of predicting the 3D poses from scratch, we first lift 2D pose history into 3D to obtain a noisy 3D human pose sequence~$\bar{{\bf Y}}_{1:N+T}$ as input, and further use~$\Psi$ to refine them to obtain the final prediction.
Our initial estimation consists of two steps.
We first obtain a noisy 3D poses~$\bar{\bf Y}_{1:N}$ by back-projecting 2D pose history~${\bf X}_{1:N}$ into 3D using the human torso depth~$\hat{\bf d}^{r}_{1:N}$ and camera intrinsics~${\bf K}$. 
We next replicate the present 3D pose~$\bar{\bf Y}_{N}$ to each of the predicted future 3D path location for an initial estimation of future 3D poses~$\bar{\bf Y}_{N+1:N+T}$.
We then concatenate both estimations together to form~$\bar{\bf Y}_{1:N+T}$ as input to our PoseNet:
\begin{equation}
    \hat{{\bf Y}}_{N+1:N+T}
    =
    \Psi(\bar{{\bf Y}}_{1:N+T}).
\end{equation}
The training objective for PoseNet is to minimize the distance between the 3D pose prediction and the ground-truth defined as:
\begin{equation}
    L_{\text{pose3D}} =
        \|{\bf Y}_{N+1:N+T} - \hat{{\bf Y}}_{N+1:N+T}\|_1.
\end{equation}
During training, ground-truth 3D path ${\bf Y}^{r}_{1:N+T}$ is used for estimating coarse 3D pose input.
During testing, we use the predicted 3D path $\hat{{\bf Y}}^{r}_{1:N+T}$ from PathNet.

%% file: hmp_3dataset.tex
\section{GTA Indoor Motion Dataset} \label{GTA}

We introduce the GTA Indoor Motion dataset (GTA-IM)\footnote[2]{Dataset available in \url{https://github.com/ZheC/GTA-IM-Dataset}} that emphasizes human-scene interactions.
Our motivation for this dataset is that
%real indoor human datasets are mostly excluded from the scenes like~\cite{IonescuSminchisescu11, straub2019replica} and CMU MoCap~\cite{CMUMocap}.
existing real datasets on human-scene interaction~\cite{hassan2019resolving,savva2016pigraphs} have relatively noisy 3D human pose annotations and limited long-range human motion limited by depth sensors.
On the other hand, existing synthetic human datasets~\cite{fabbri2018learning,krahenbuhl2018free} focus on the task of human pose estimation or parts segmentation and sample data in wide-open outdoor scenes with limited interactable objects. %These synthetic datasets are not suitable for our problem.
More detailed comparison to these datasets is in Appendix~\ref{sec:supp-dataset}.

To overcome the above issues, we spend extensive efforts in collecting a synthetic dataset by developing an interface with the game engine for controlling characters, cameras, and action tasks in a fully automatic manner.
For each character, we randomize the goal destination inside the 3D scene, the specific task to do, the walking style, and the movement speed.
We control the lighting condition by changing different weather conditions and daytime.
We also diversify the camera location and viewing angle over a sphere around the actor such that it points towards the actor. We use in-game ray tracing API and synchronized human segmentation map to track actors.
%When the actor walks outside the field of view, the camera will be resampled immediately.
%When the actor finishes all assigned tasks, we stop recording and start a new sequence.
The collected actions include climbing the stairs, lying down, sitting, opening the door, and etc. -- a set of basic activities within indoor scenes. For example, the character has 22 walking styles including 10 female and 12 male walking styles. All of these factors enable us to collect a diverse and realistic dataset with accurate annotations for our challenging task.

\begin{figure}[t]
    \centering
    \includegraphics[width=\textwidth]{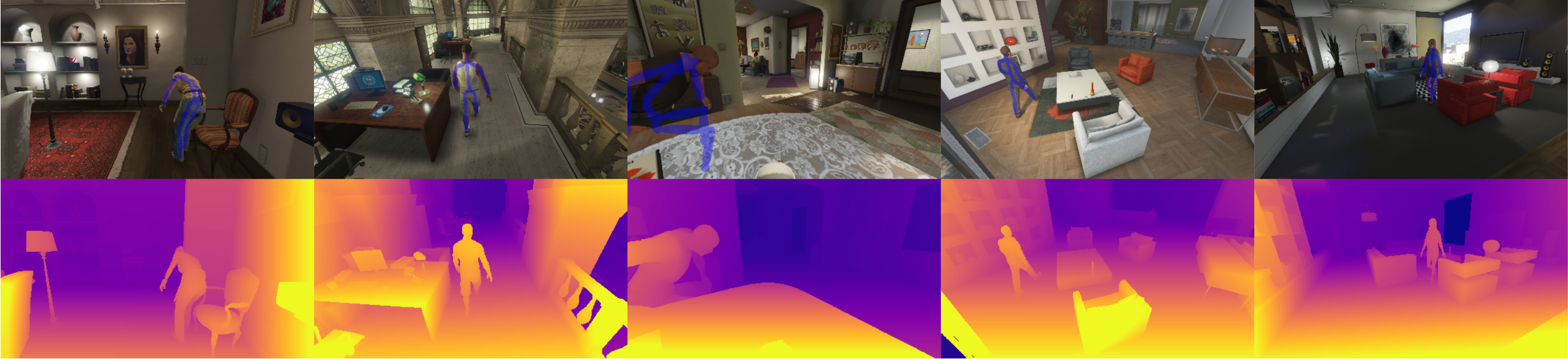} \\
    \vspace{-5pt}
    \caption{\small {\bf Sample RGBD images from GTA-IM dataset.} Our dataset contains realistic RGB images (visualized with the 2D pose), accurate depth maps, and clean 3D human pose annotations. 
    }
    \figlabel{gta}
    \vspace{-15pt}
\end{figure}

In total, we collect one million RGBD frames of $1920\times1080$ resolution with the ground-truth 3D human pose (98 joints), human segmentation, and camera pose.
Some examples are shown in~\figref{gta}. The dataset contains $50$ human characters acting inside $10$ different large indoor scenes. Each scene has several floors, including living rooms, bedrooms, kitchens, balconies, and etc., enabling diverse interaction activities.

%% file: hmp_4experiments.tex
\section{Evaluation}\label{sec:exp}
We perform extensive quantitative and qualitative evaluations of our future 3D human path and motion predictions.
The rest of this section is organized as follows:
We first describe the datasets we use in Section~\ref{datasets}.
We then elaborate on our quantitative evaluation metrics and strong baselines in Section~\ref{baseline}.
Further, we show our quantitative and qualitative improvement over previous methods in Section~\ref{quantitative}.
Finally, we evaluate our long-term  predictions and show qualitative results of destination samples and final 3D pose results in Section~\ref{sec:stochastic}. We discussed some failure cases in Section~\ref{sec:failure}. Implementation details are described in Appendix~\ref{sec:details}.

\subsection{Datasets}
\label{datasets}
\vspace{2pt} \boldstart{GTA-IM:} We train and test our model on our collected dataset as described in Section~\ref{GTA}. We split $8$ scenes for training and $2$ scenes for evaluation. We choose $21$ out of $98$ human joints provided from the dataset. We convert both the 3D path and the 3D pose into the camera coordinate frame for both training and evaluation.

\vspace{2pt} \boldstart{PROX:}  Proximal Relationships with Object eXclusion (PROX) is a new dataset captured using the Kinect-One sensor by Hassan et al.~\cite{hassan2019resolving}. It comprises of 12 different 3D scenes and RGB sequences of 20 subjects moving in and interacting with the scenes. We split the dataset with 52 training sequences and 8 sequences for testing. Also, we extract 18 joints from the SMPL-X model~\cite{SMPL-X:2019} from the provided human pose parameters.

\subsection{Evaluation Metric and Baselines}\label{baseline}
\vspace{2pt} \boldstart{Metrics:} 
We use the Mean Per Joint Position Error (MPJPE)~\cite{ionescu2013human3} as a metric for quantitatively evaluating both the 3D path and 3D pose prediction. We report the performance at different time steps (seconds) in millimeters (mm). 

\vspace{2pt} \boldstart{Baselines:} To the best of our knowledge, there exists no prior work that predicts 3D human pose with global movement using 2D pose sequence as input. Thus, we propose three strong baselines for comparison with our method. For the first baseline, we combine the recent 2D-to-3D human pose estimation method~\cite{pavllo20193d} and 3D human pose prediction method~\cite{wei2019motion}.
For the second baseline, we use Transformer~\cite{vaswani2017attention}, the state-of-the-art sequence-to-sequence model, to perform 3D prediction directly from 2D inputs treating the entire problem as a single-stage sequence to sequence task. For the third baseline, we compare with is constructed by first predicting the future 2D human pose using~\cite{vaswani2017attention} from inputs and then lifting the predicted pose into 3D using~\cite{pavllo20193d}.
Note that none of these baselines consider scene context or deal with uncertainty in their future predictions. We train all models on both datasets for two-second-long prediction based on 1-second-long history and report their best performance for comparison.

\subsection{Comparison with Baselines}
\label{quantitative}
In this section, we perform quantitative evaluations of our method in the two datasets. We also show some qualitative comparisons in~\figref{qualitative_comparison}. We evaluate the two modes of our model: the stochastic mode that can generate multiple future predictions by sampling different 2D destinations from the GoalNet; and the deterministic mode that can generate one identical prediction without deploying GoalNet.%, i.e., removing the destination heatmap from the input when training and testing PathNet. As a result, the model generates deterministic output from the same input. 

\begin{table}[h]
    \centering
    \resizebox{0.7\textwidth}{!}{\
    \begin{tabular}{lccccccccc}
        \toprule
        & \multicolumn{4}{c}{3D path error (mm)} & \multicolumn{5}{c}{3D pose error (mm)}  \\
        \cmidrule(lr){2-5} \cmidrule(lr){6-10}
        Time step (second) & 0.5 & 1 & 1.5 & 2 & 0.5 & 1 & 1.5 & 2 & \textbf{All} $\downarrow$\\
        \midrule
        TR~\cite{vaswani2017attention}  & 277 & 352 & 457 & 603 & 291&    374&    489&    641 & 406\\
        TR~\cite{vaswani2017attention} + VP~\cite{pavllo20193d}  & 157 & 240 & 358 & 494 & 174 & 267 &388 & 526 & 211 \\
        VP~\cite{pavllo20193d} + LTD~\cite{wei2019motion} & 124 &    194 &    276    & 367 & 121&    180&    249&    330 &193\\
        \midrule
        Ours (deterministic) &
        \textbf{104} &
        \textbf{163} &
        \textbf{219} &
        \textbf{297} &
        \textbf{91} &
        \textbf{158} &
        \textbf{237} &
        \textbf{328} &
        \textbf{173}
        \\
        \midrule
        Ours (samples=4) &114 &162 &227 &310 &94 &161 &236 &323 &173\\
        Ours (samples=10) &110 &154 &213 &289 &90 &154 &224 &306 &165\\
        \midrule
        Ours w/ xyz. output &122 &179 &252 &336 &101 &177 &262 &359 &191\\
        Ours w/o image &128 &177 &242 &320 &99 &179 &271 &367 &196\\
        Ours w/ masked image &120 &168 &235 &314 &96 &170 &265 &358 &189 \\
        Ours w/ RGBD input &100 &138 &193 &262 &93 &160 &235 &322 &172 \\
        Ours w/ GT destination &104 &125 &146 &170 &85 &133 &178 &234 &137\\
        \bottomrule
    \end{tabular}
    }
    \vspace{5pt}
    \caption{\small
        {\bf Evaluation results in GTA-IM dataset}. We compare other baselines in terms of 3D path and pose error. The last column (All) is the mean average error of the entire prediction over all time steps. VP denotes Pavllo et al.~\cite{pavllo20193d}, TR denotes transformer~\cite{vaswani2017attention} and LTD denotes Wei et al.~\cite{wei2019motion}.
        GT stands for ground-truth, xyz. stands for directly regressing 3D coordinates of the path. We report the error of our stochastic predictions with varying number of samples. 
    }
    \label{tab:gta_all}
    \vspace{-15pt}
\end{table}

\vspace{2pt} \boldstart{GTA-IM:}
The quantitative evaluation of 3D path and 3D pose prediction in GTA-IM dataset is shown in Table~\ref{tab:gta_all}.
%As the trends are similar on 3D path and pose prediction, we refer to final 3D pose results in our analysis.
Our deterministic model with image input can outperform the other methods by a margin, i.e., with an average error of $173$ mm vs. $193$ mm. When using sampling during inference, the method can generate multiple hypotheses of the future 3D pose sequence. We evaluate different numbers of samples and select the predictions among all samples that best matches ground truth to report the error. We find using four samples during inference can match the performance of our deterministic model ($173$ mm error), while using ten samples, we further cut down the error to $165$ mm. These results validate that our stochastic model can help deal with the uncertainty of future human motion and outperform the deterministic baselines with few samples.

As an ablation, we directly regress 3D coordinates (``Ours w/ xyz.'' in the Table~\ref{tab:gta_all}) and observe an overall error that is $18$ mm higher than the error of our deterministic model ($191$ mm vs. $173$ mm). This validates that representing the 3D path as the depth and 2D heatmap of the human center is better due to its strong correlation to the image appearance. We also ablates different types of input to our model. Without image input, the average error is $23$ mm higher. With only masked images input, i.e., replacing pixels outside human crop by ImageNet mean pixel values, the error is $16$ mm highe. This validates that using full image to encode scene context is more helpful than only observing cropped human image, especially for long-term prediction. Using both color and depth image as input (``Ours w/ RGBD input''), the average error is $172$ mm which is similar to the model with RGB input. This indicates that our model implicitly learns to reason about depth information from 2D input. 
If we use ground-truth 2D destinations instead of predicted ones, and the overall error decreases down to $137$ mm. It implies that the uncertainty of the future destination is the major source of difficulty in this problem.

\begin{table}[h]
    \vspace{-5pt}
    \centering
    \resizebox{0.75\textwidth}{!}{\
    \begin{tabular}{lccccccccc}
        \toprule
        & \multicolumn{4}{c}{3D path error (mm)} & \multicolumn{5}{c}{3D pose error (mm)}  \\
        \cmidrule(lr){2-5} \cmidrule(lr){6-10}
        Time step (second) & 0.5 & 1 & 1.5 & 2 & 0.5 & 1 & 1.5 & 2 & \textbf{All} $\downarrow$\\
        \midrule
        TR~\cite{vaswani2017attention}  & 487 & 583 &    682&    783 & 512 & 603 & 698 & 801 & 615 \\
        TR~\cite{vaswani2017attention} + VP~\cite{pavllo20193d}  & 262 & 358 & 461 & 548 &297 & 398 & 502 & 590 & 326\\
        VP~\cite{pavllo20193d} + LTD~\cite{wei2019motion} &
        194    &
        263    &
        332    &
        394 &
        216 &
        274 &
        335 &
        \textbf{394} &
        282
        \\
        \midrule
        Ours w/o GTA-IM pretrain &192 &258 &336 &419 &192 &273 &352 &426 &280\\
        Ours (deterministic) &
        \textbf{189} &
        \textbf{245} &
        \textbf{317} &
        \textbf{389} &
        \textbf{190} &
        \textbf{264} &
        \textbf{335} &
        406 &
        \textbf{270} \\
        \midrule
        Ours (samples=3) &192 &245 &311 &398 &187 &258 &328 &397 & 264 \\
        Ours (samples=6) &185 &229 &285 &368 &184 &249 &312 &377 &254\\
        Ours (samples=10) &181 &222 &273 &354&182 &244 &304 &367 &249 \\
        \midrule
        Ours w/ gt destination &193 &223 &234 &237 &195 &235 &276 &321 &237 \\
        \bottomrule
    \end{tabular}
    }
    \vspace{5pt}
    \caption{\small
        {\bf
            Evaluation results in PROX dataset}. We compare other baselines in terms of 3D future path and pose prediction. VP denotes Pavllo et al.~\cite{pavllo20193d}, TR denotes transformer~\cite{vaswani2017attention} and LTD denotes Wei et al.~\cite{wei2019motion}. GT stands for ground-truth. We rank all methods using mean average error of the entire prediction (last column).
    }
    \label{tab:prox_all}
    \vspace{-20pt}
\end{table}

\vspace{4pt} \boldstart{PROX:} The evaluation results in Table~\ref{tab:prox_all} demonstrate that our method outperforms the previous state of the art in terms of mean MPJPE of all time steps, $270$ mm vs. $282$ mm. Overall, we share the same conclusion as the comparisons in GTA-IM dataset. When using sampling during inference, three samples during inference can beat the performance of our deterministic model ($264$ mm vs. $270$ mm), while using ten samples, the error decreases to $249$ mm. Note that these improvements are more prominent than what we observe on GTA-IM benchmark. This is because the uncertainty of future motion in the real dataset is larger. Therefore, stochastic predictions have more advantage.

Moreover, we find that pre-training in GTA-IM dataset can achieve better performance ($270$ mm vs. $280$ mm). Our method exploits the image context and tends to overfit in PROX dataset because it is less diverse in terms of camera poses and background appearance (both are constant per video sequence). Pre-training in our synthetic dataset with diverse appearance and clean annotations can help prevent overfitting and boost the final performance. %Note the error of the best model in PROX ($270$ mm) is still relatively high compared to that in GTA-IM of ($173$ mm). The main reason we observe is that PROX dataset has relatively noisy 3D human poses, i.e, temporal jittering, due to the difficulty of obtaining accurate 3D poses in the real-world setting. %As a result, the model supervised by relatively noisy pseudo ground truth has difficulty in stable training and evaluation.
%Our synthetic dataset with diverse appearance and clean annotations is helpful for stable training and rigorous evaluation.

\begin{figure}[h]
 \vspace{-15pt}
    \centering
    \includegraphics[width=1.0\textwidth]{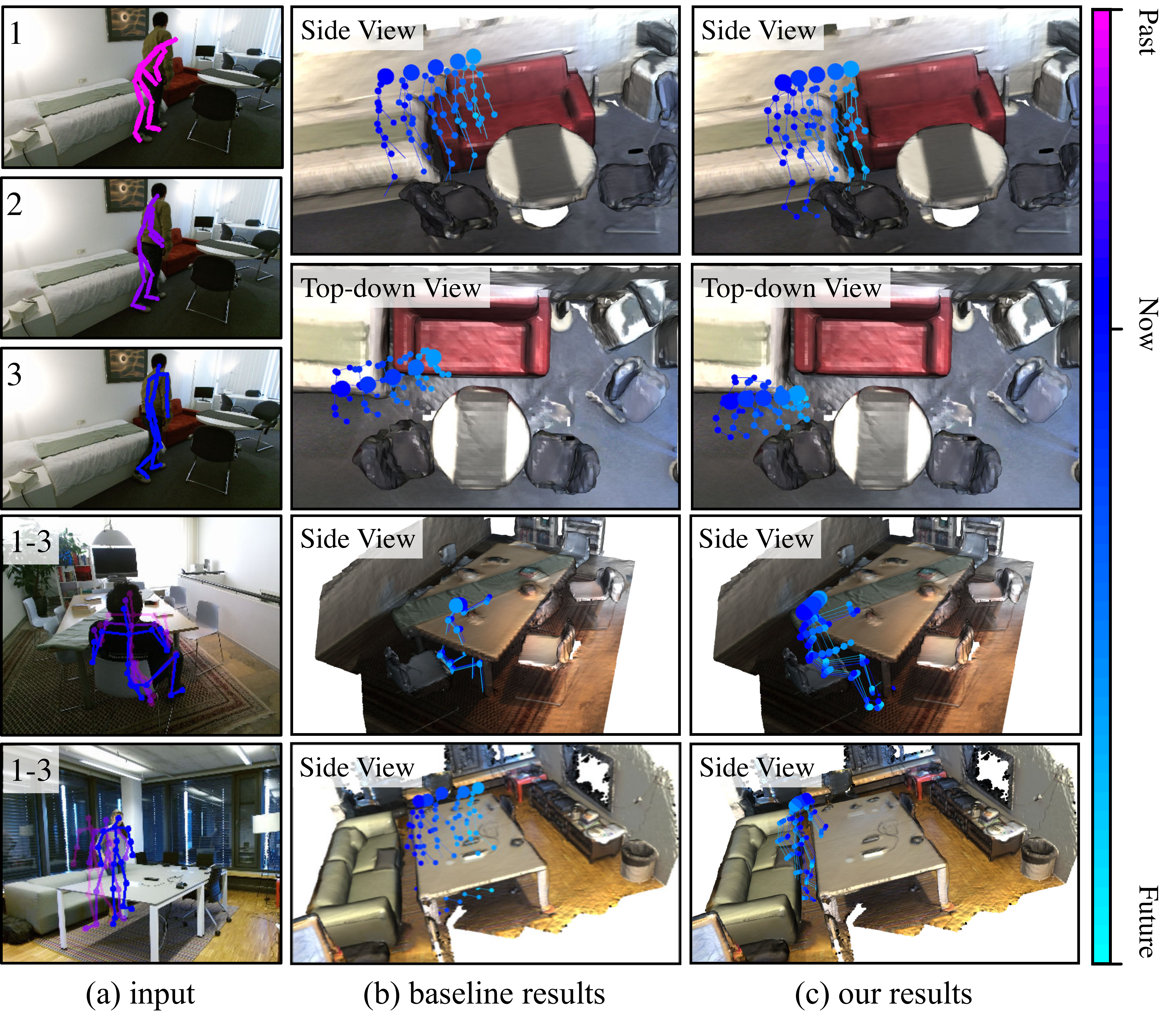} \\
    \vspace{-10pt}
    \caption{\small \textbf{Qualitative comparison}. We visualize the input (a), the results of VP\cite{pavllo20193d} and LTD~\cite{wei2019motion} (b) and our results (c) in the ground-truth 3D mesh. The color of pose is changed over timesteps according to the color bar. The first example includes both top-down and side view. From the visualization, we can observe some collisions between the baseline results and the 3D scene, while our predicted motion are more plausible by taking the scene context into consideration.
    }
    \figlabel{qualitative_comparison}
    \vspace{-15pt}
\end{figure}

\vspace{2pt} \boldstart{Qualitative comparison:} In~\figref{qualitative_comparison}, we show qualitative comparison with the baseline of VP~\cite{pavllo20193d} and LTD~\cite{wei2019motion}. This baseline is quantitatively competitive as shown in Table~\ref{tab:gta_all} and~\ref{tab:prox_all}. However, without considering scene context, their predicted results may not be feasible inside the 3D scene, e.g., the person cannot go through a table or sofa. 
In contrast, our model implicitly learns the scene constraints from the image and can generate more plausible 3D human motion in practice.

\begin{figure*}[t]
    \centering
    \includegraphics[width=0.8\linewidth]{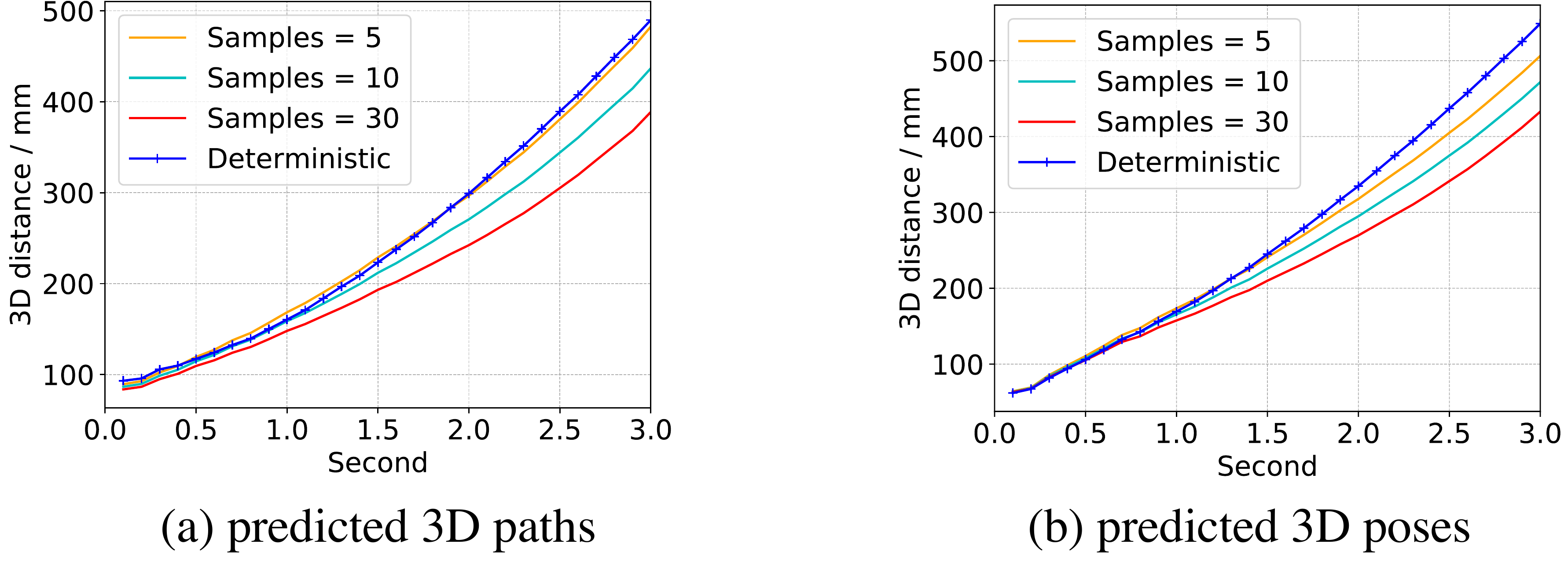}
    \vspace{-6pt}
    \caption{\small
        \textbf{
            Comparison between our stochastic predictions and deterministic predictions.
        }
        We show error curves of predicted (a) 3D paths and (b) 3D poses with varying numbers of samples over varying timesteps on GTA-IM dataset. In all plots, we find that our stochastic model can achieve better results with a small number of samples, especially in the long-term prediction (within 2-3 second time span).
    }
    \label{fig:deter-vs-multi}
    %\vspace{-10pt}
\end{figure*}

\begin{figure}[t]
    \centering
    \includegraphics[width=1.0\textwidth]{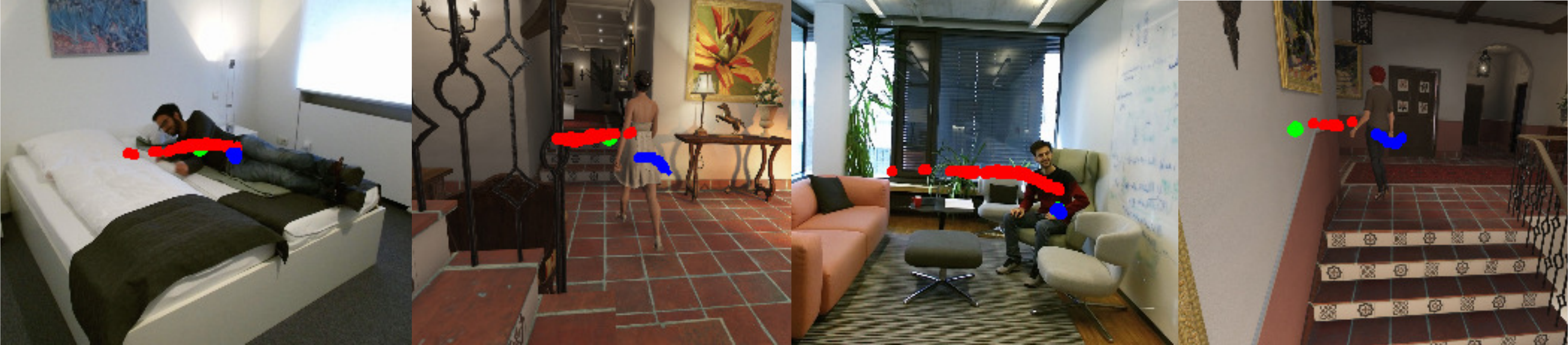} \\
    \vspace{-5pt}
    \caption{\small \textbf{Destination sampling results}. In each image, the blue dots denote the path history, the green dots are ground-truth future destination, red dots are sample destinations from the GoalNet which we draw 30 times from the standard Gaussian. Our method can generate diverse plausible motion destination samples which leads to different activities, i.e., sitting still or standing up.
    }
    \figlabel{destinations}
    \vspace{-10pt}
\end{figure}

\subsection{Evaluation and Visualization on Longer-term Predictions}
\label{sec:stochastic}
To demonstrate our method can predict future human motion for more than 2 seconds, we train another model to produce the 3-second-long future prediction.
In Figure~\ref{fig:deter-vs-multi}, we show the self-comparisons between our stochastic predictions and deterministic predictions.
Our stochastic models can beat their deterministic counterpart using 5 samples. With increasing number of samples, the testing error decreases accordingly. The error gap between deterministic results and stochastic results becomes larger at the later stage of the prediction, i.e., more than $100$ mm difference at $3$ second time step. This indicates the advantage of the stochastic model in long-term future motion prediction.

We show qualitative results of our stochastic predictions on movement destinations in~\figref{destinations}, and long-term future motion in~\figref{qualitative}. Our method can generate diverse human movement destination, and realistic 3D future human motion by considering the environment, e.g., turning left/right, walking straight, taking a U-turn, climbing stairs, standing up from sitting, and laying back on the sofa.

\begin{figure*}[!h]
    \centering
    \includegraphics[width=0.95\textwidth]{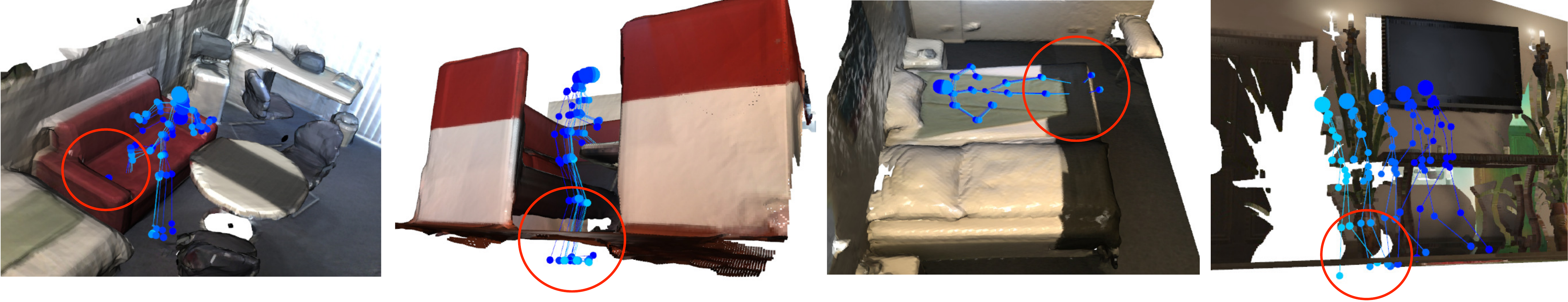} \\
    \vspace{-5pt}
    \caption{\small \textbf{Visualization of failure cases}. In each red circle area, we observe the intersection between the human feet and the 3D mesh, e.g., the bed.
    }
    \vspace{-15pt}
    \figlabel{failure}
\end{figure*}

\subsection{Discussion of failure cases} \label{sec:failure}
Our model implicitly learns scene constraints in a data-driven manner from large amounts of training data, and can produce consistent 3D human paths without serious collision comparing to previous methods which do not take scene context into consideration as shown in \figref{qualitative_comparison}. However, without assuming we have access to the pre-reconstructed 3D mesh and using expensive offline optimization as~\cite{hassan2019resolving}, the resulting 3D poses may not strictly meet all physical constraints of the 3D scene geometry. Some failure cases are shown in \figref{failure}. In the red circled area, we observe small intersections between the human feet and the 3D scene mesh, e.g., the ground floor or the bed. This issue could be relieved by integrating multi-view/temporal images as input to the learning system to recover the 3D scene better. The resulting 3D scene could be further used to enforce explicit scene geometry constraints for refining the 3D poses. We leave this to the future work.

%% file: hmp_5conclusions.tex
\section{Conclusion}
In this work, we study the challenging task of long-term 3D human motion prediction with only 2D input. This research problem is very relevant to many real-world applications where understanding and predicting feasible human motion considering the surrounding space is critical, e.g., a home service robot collaborating with the moving people, AR glass providing navigation guide to visually impaired people, and autonomous vehicle planning the action considering the safety of pedestrians. We present an initial attempt in attacking the problem by contributing a new dataset with diverse human-scene interactions and clean 3D annotations. We also propose the first method that can predict long-term stochastic 3D future human motion from 2D inputs, while taking the scene context into consideration. There are still many aspects in this problem that can be explored in the future, such as how to effectively evaluate the naturalness and feasibility of the stochastic human motion predictions, and how to incorporate information of dynamic objects and multiple moving people inside the scene.

\vspace{8pt}
\boldstart{Acknowledgements} We thank Carsten Stoll and Christoph Lassner for their insightful feedback. We are also very grateful to the feedback from the BAIR community members including but not limited to Angjoo Kanazawa, Xiaolong Wang, Ashish Kumar, Allan Jabri, Ilija Radosavovic, Sasha Sax, Lerrel Pinto, and Deepak Pathak.
%We thank FRL Sausalito office for the support and the helpful feedback from the team researchers including but not limited to Carsten Stoll and Christoph Lassner. We are also very grateful to the feedback from the BAIR community members including but not limited to Angjoo Kanazawa, Xiaolong Wang, Ashish Kumar, Allan Jabri, Ilija Radosavovic, Sasha Sax, Lerrel Pinto, and Deepak Pathak.

% \newpage

%% file: main_supplement.tex
\section*{APPENDIX}
\vspace{-0.6em}
In the supplement, we first compare relevant datasets in Section~\ref{sec:supp-dataset}. We also cover the implementation details in Section~\ref{sec:details}. Further, we evaluate GoalNet in Section~\ref{sec:supp-goalnet} and long-term predictions on PROX in Section~\ref{sec:supp-prox-long}. We also show qualitative comparison results in Section~\ref{sec:supp-qualitative}. 
Finally, we describe the network architecture in details in Section~\ref{sec:supp-network}. We include more visual results on 3D path and 3D pose prediction in our {\bf video} which can be found at \url{https://people.eecs.berkeley.edu/~zhecao/hmp/}.

\section{Dataset Comparison} \label{sec:supp-dataset}
\vspace{-15pt}
\begin{table}[!h]
\scriptsize
\begin{center}
\setlength{\tabcolsep}{4pt}
\label{table:datasets}
\begin{tabular}{lllccccccc}
\hline\noalign{\smallskip}
Dataset             & \#Clips  & \#Frames     & \#View per clip    & \#Characters        &\#Scenes    & Pose jittering  & Depth Range & 3D scene\\
\noalign{\smallskip}
\hline
\noalign{\smallskip}
H3.6M \cite{ionescu2013human3} & 80 & 3.6 M & 4 & 11 (6M 5F) & 1 & & 0 -- 6&\\
PiGraph \cite{savva2016pigraphs}    & 63         & 0.1 M     &      1    &    5 (4M 1F)    & 30    & \checkmark  & 0 -- 4 & \checkmark\\
PROX \cite{hassan2019resolving}            & 60        & 0.1 M     &  1    &    20 (16M 4F)    &  12    &     \checkmark    &0 -- 4 & \checkmark\\
\noalign{\smallskip}
\hline
\noalign{\smallskip}
Ours & 119 & 1 M &  14-67 & 50 (25M 25F) & 49 &   &0 -- 200 & \checkmark\\
\hline
\end{tabular}
\vspace{6pt}
\caption{\small Overview of the publicly available datasets on 3D human motion.}
\label{supp:dataset}
\end{center}
\vspace{-30pt}
\end{table}

In Table~\ref{supp:dataset}, we list some representative datasets on 3D human motion including Human3.6M (H3.6M)~\cite{ionescu2013human3}, PiGraphs~\cite{savva2016pigraphs}, and Proximal Relationships with Object eXclusion (PROX)~\cite{hassan2019resolving}. H3.6M~\cite{ionescu2013human3} is a large-scale dataset with accurate 3D pose annotations using a multi-camera capturing system.
However, all recordings were captured in the lab environment (mostly empty space) and thus it lacks diverse human interaction with the indoor environment, e.g., sitting on a sofa or climbing stairs.
PiGraphs~\cite{savva2016pigraphs} and PROX~\cite{hassan2019resolving}, on the other hand, are dedicated datasets with extensive efforts for modeling the interaction between 3D human and 3D scene.
Due to extensive efforts required for manually collecting the RGBD sequence of human activities, both datasets have a relatively small number of frames, scenes, and characters.
They are also less diverse in terms of camera poses and background appearance (only one static camera viewpoint for each entire video clip).
As shown in our experiments (Table 1 in the main paper), models trained on these datasets tend to be overfitting to the training data. 
Their 3D human poses are also relatively noisy, e.g., temporal jittering, due to the difficulty of obtaining accurate 3D poses in the real-world setting.

In contrast, we collect a large-scale and more diverse dataset with clean annotations by developing an automatic data collection pipeline based on the gaming engine. We diversify the camera location and viewpoint over a sphere around the actor such that it points towards the actor. We use in-game ray tracing API and synchronized human segmentation map to track actors. When the actor walks outside the field of view, the camera will be resampled immediately. We believe our synthetic dataset with clean annotations can be complementary to real data for stable training and rigorous evaluation.

\section{Implementation Details}
\label{sec:details}
In this section, we describe the implementation details for each module of the model. All modules are implemented in PyTorch~\cite{paszke2017automatic} and trained using the ADAM optimizer~\cite{kingma2014adam}. We set the input image size and the heatmap size to 256$\times$448; the resolution of output future heatmap prediction to 64$\times$112; all depth dimension values are caped by 10 and normalized by a factor 4 during training. We train all 3 modules separately and find it works better than joint training. The multi-modal nature of this problem makes it hard to train the model with intermediate prediction, e.g., it is not quite reasonable to supervise PathNet with ground-truth (GT) 3D path towards the GT destination, when taking the input of a very different destination predicted by GoalNet. 

\vspace{2pt} \boldstart{GoalNet:}
We use a $10^{-4}$ learning rate without weight decay. For both datasets, we train for 2 epochs with a batch size of 128.

\vspace{2pt} \boldstart{PathNet:}
We train our PathNet with ground-truth destination input, while during inference, we use the prediction from GoalNet instead.
The learning rate is set to $2.5\times10^{-4}$ with a $10^{-4}$ weight decay. Our models are trained for $10$ epochs for GTA-IM and $6$ epochs for PROX where learning rates decay by a factor of 10 at $7$th and $4$th epochs, respectively. We use a batch size of 32.

\vspace{2pt} \boldstart{PoseNet:}
We train the PoseNet ground-truth 3D path, while during inference, we use the prediction from PoseNet instead. We train the model for $80$ epochs using a learning rate of $1\times10^{-3}$, an attention dropout rate of $0.2$, and batch size $1024$.

\section{GoalNet Evaluation} \label{sec:supp-goalnet}
\begin{table}[h]
\vspace{-23pt}
    \centering
    \resizebox{0.65\textwidth}{!}{\
    \begin{tabular}{lcccc}
        \toprule
        & \multicolumn{2}{c}{GTA-IM dataset} & \multicolumn{2}{c}{PROX dataset}  \\
        \cmidrule(lr){2-3} \cmidrule(lr){4-5}
        Dataset &
        min &
        avg{\tiny $\pm$std} &
        min &
        avg{\tiny $\pm$std} \\
        \midrule
        Ours (deterministic) &
        23.7 &
        - &
        27.7 &
        - \\
        \midrule
        Ours (samples=3) &
        25.3 &
        41.6{\tiny $\pm10.3$} &
        30.3 &
        38.3{\tiny $\pm7.1$}
        \\
        Ours (samples=5) &
        23.6 &
        40.3{\tiny $\pm12.7$} &
        27.7 &
        37.2{\tiny $\pm8.3$}
        \\
        Ours (samples=10) &
        17.6 &
        34.6{\tiny $\pm14.0$} &
        24.9 &
        35.3{\tiny $\pm9.1$}
        \\
        Ours (samples=30) &
        \textbf{12.2} &
        35.4{\tiny $\pm17.2$} &
        \textbf{21.7} &
        31.3{\tiny $\pm9.7$}
        \\
        \bottomrule
    \end{tabular}
    }
    \vspace{10pt}
    \caption{\small
        {\bf
            Evaluation of 2D goal predictions in GTA-IM and PROX dataset}.
            We compare our results of directly predicting 2D destination using GoalNet with those obtained by our deterministic PathNet.
            We compare them in terms of the least and average error among all samples.
    }
    \label{tab:2d_goal}
    \vspace{-15pt}
\end{table}

In Table~\ref{tab:2d_goal}, we evaluate 2D future destination predictions of GoalNet.
We use the metric of Mean Per Joint Position Error (MPJPE)~\cite{ionescu2013human3} in the 2D image space.
We compare the stochastic results sampled from GoalNet with the deterministic results.
We vary the number of samples during the evaluation and present results on both datasets. Our findings are twofold.
(1) Directly predicting 2D destinations is beneficial.
Our GoalNet can achieve similar performance with the deterministic baseline using as few as 5 samples on both datasets.
(2) With more samples, our prediction performance increases monotonously.
When using 30 samples, our GoalNet can outperform the deterministic baseline by large margins, bringing around 40\% less error on GTA-IM dataset and 20\% less error on PROX dataset.

\section{Long-term Evaluation on PROX} \label{sec:supp-prox-long}
\begin{figure*}[t]
    \centering
    \begin{subfigure}[b]{0.45\textwidth}
        \includegraphics[width=\linewidth]{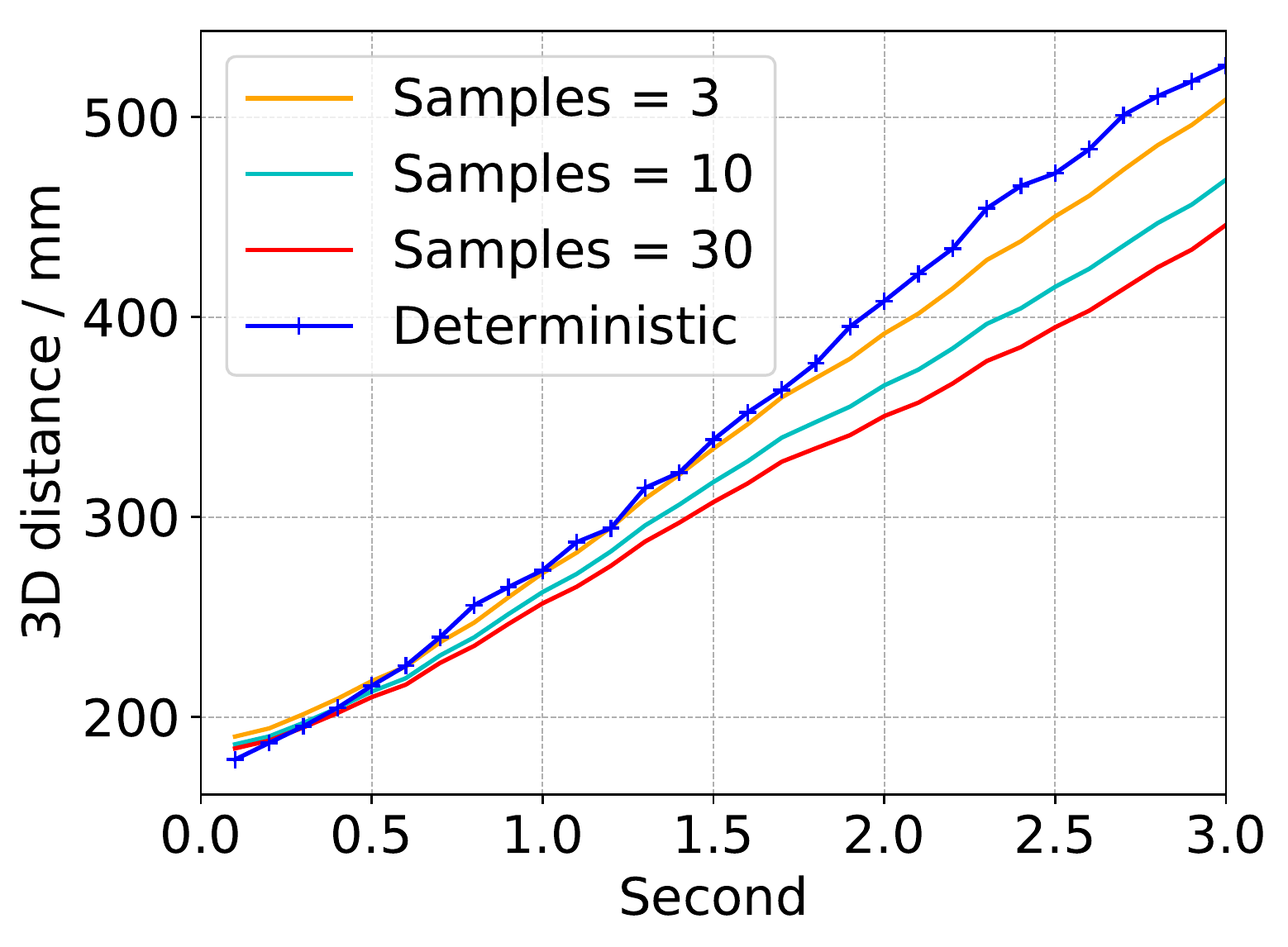}
        \caption{predicted 3D paths}
    \end{subfigure}%
    \begin{subfigure}[b]{0.45\textwidth}
        \includegraphics[width=\linewidth]{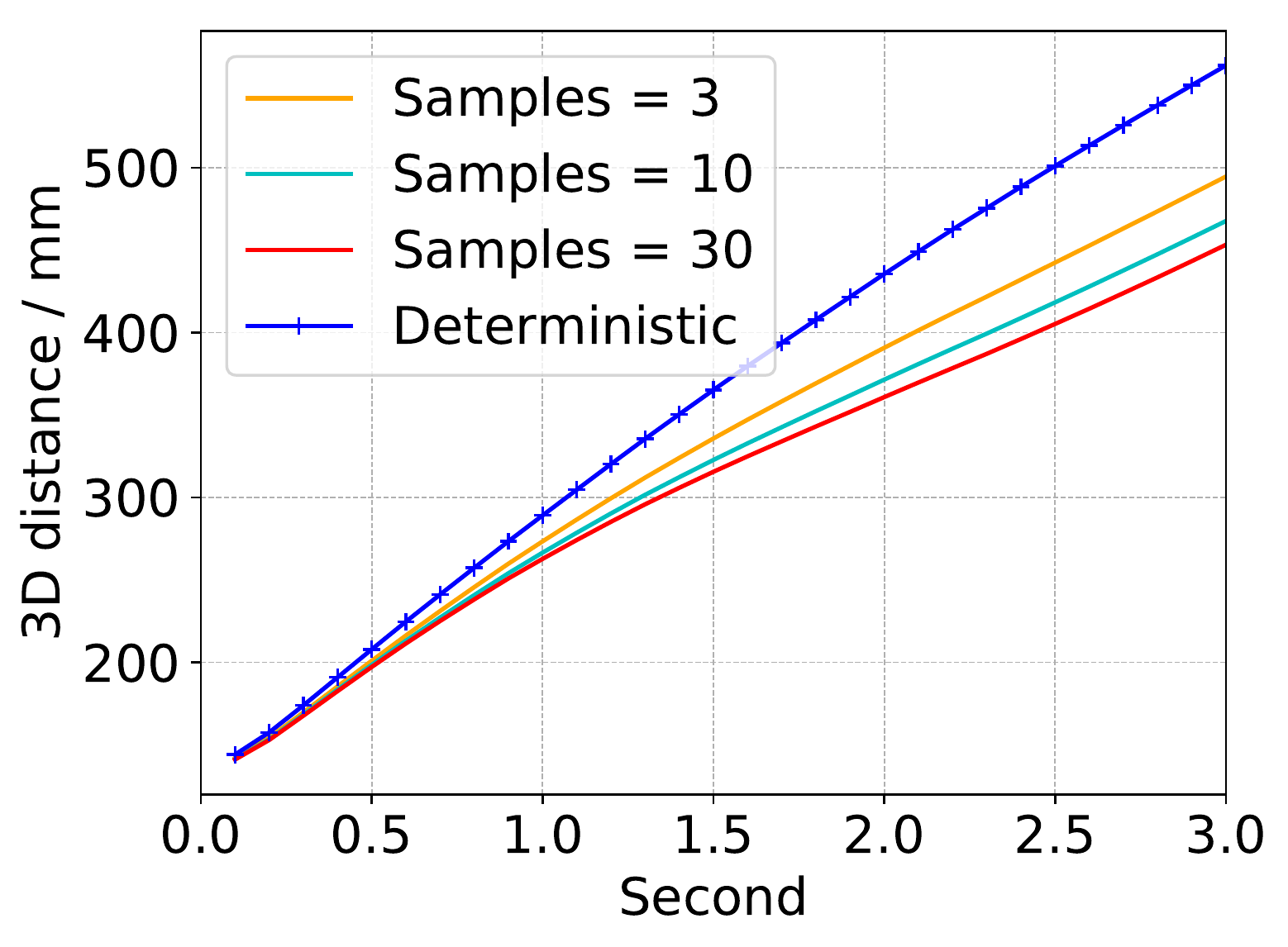}
        \caption{predicted 3D poses}
    \end{subfigure}%
    \vspace{-6pt}
    \caption{\small
        \textbf{
            Comparison between our stochastic predictions and deterministic predictions of long-term prediction on PROX.
        }
        We show error curves of predicted (a) 3D paths and (b) 3D poses with varying numbers of samples over varying timesteps on GTA-IM dataset. In all plots, we find that our stochastic model can achieve better results with a small number of samples, especially in the long-term prediction (within 2-3 seconds time span).
    }
    \label{fig:supp-deter-vs-multi}
    %\vspace{-10pt}
\end{figure*}

In Figure~\ref{fig:supp-deter-vs-multi}, we evaluate long-term prediction results on PROX dataset as we previously showed on GTA-IM dataset.
Specifically, we show results on predicted 3D paths and predicted 3D poses using our deterministic model and stochastic model with varying numbers of samples.
We note similar trends as previously seen on GTA-IM dataset.
More interestingly, we find that our stochastic models can beat their deterministic counterpart using only 3 samples on PROX, compared to 5 samples on GTA-IM.

\section{Qualitative Results} \label{sec:supp-qualitative}

We show additional qualitative comparison results in \figref{qualitative_comp2}.
More results on 3D path and 3D pose prediction are in our {\bf video} which can be found at \url{https://people.eecs.berkeley.edu/~zhecao/hmp/}.

\section{Network Architecture} \label{sec:supp-network}
We outline our network architectures in this section.
Specifically, we define our GoalNet in Table~\ref{tab:supp-goalnet}, PathNet in Table~\ref{tab:supp-pathnet}. For PoseNet, please refer to~\cite{vaswani2017attention}, we modified the architecture by removing the input embedding layer, output embedding layer, positional encoding layer and the softmax layer.

\begin{figure*}[!h]
    \centering
    \includegraphics[width=1.0\textwidth]{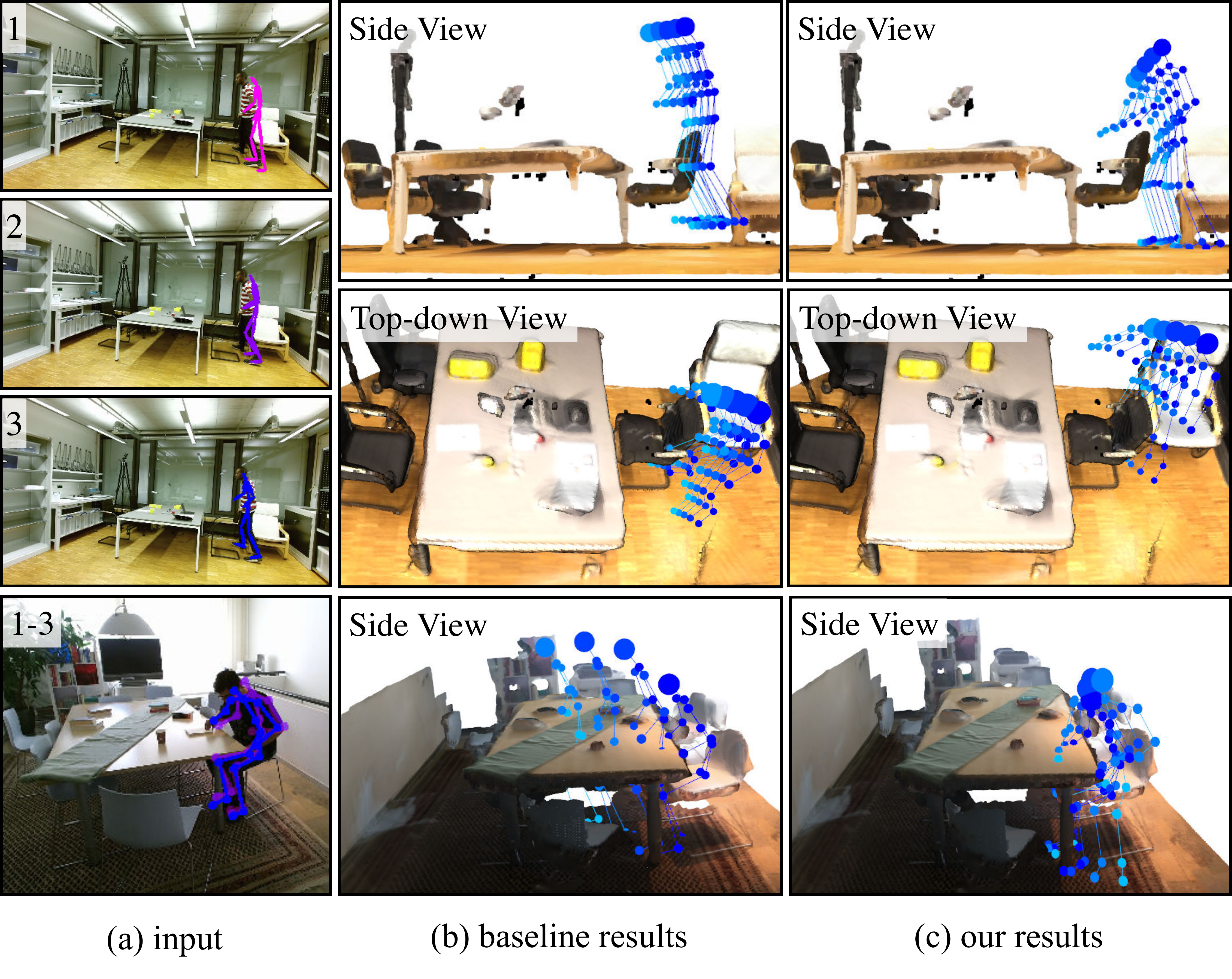} \\
    \caption{\small \textbf{Qualitative comparison}. We visualize the input (a), the results of VP\cite{pavllo20193d} and LTD~\cite{wei2019motion} (b) and our results (c) in the ground-truth 3D mesh. The color of pose is changed over timesteps from purple to dark blue and finally light blue. The first example (the 1st and 2nd row) includes both top-down view and side view of the results. From the visualization, we can observe some collisions between the baseline results and the 3D scene, while our predicted motion are more plausible by taking the scene context into consideration.
    }
    \vspace{-5pt}
    \figlabel{qualitative_comp2}
\end{figure*}

\begin{table}
  \centering
  \begin{tabular}{ccccc}
    \toprule
    Index & Input & Data & Operator & Output shape \\
    \midrule
    (1) & - & scene image & - & $3\times256\times448$ \\
    (2) & - & stacked heatmaps & - & $(N \times J)\times64\times112$ \\
    \midrule
    (3) & (1) & & $7\times7$, stride 2 & $64\times128\times224$ \\
    (4) & (3) & & MaxPool, stride 2 & $64\times64\times112$ \\
    (5) & (4) & &
    $\begin{bmatrix}
    3 \times 3, \text{stride 1} \\
    3 \times 3, \text{stride 1}
    \end{bmatrix}$
    & $64\times64\times112$ \\
    (6) & (2) & &
    $\begin{bmatrix}
    3 \times 3, \text{stride 1} \\
    3 \times 3, \text{stride 1}
    \end{bmatrix}$
    & $64\times64\times112$ \\
    (7) & (5), (6) & &
    $\begin{bmatrix}
    3 \times 3, \text{stride 2} \\
    3 \times 3, \text{stride 1}
    \end{bmatrix}$
    & $128\times32\times56$ \\
    (8) & (7) & &
    $\begin{bmatrix}
    3 \times 3, \text{stride 2} \\
    3 \times 3, \text{stride 1}
    \end{bmatrix}$
    & $256\times16\times28$ \\
    (9) & (8) & &
    $\begin{bmatrix}
    3 \times 3, \text{stride 2} \\
    3 \times 3, \text{stride 1}
    \end{bmatrix}$
    & $512\times8\times14$ \\
    (10) & (9) & encoder feat. & GlobalAvgPool & $512\times1\times1$ \\
    \midrule
    (11) & (10) & $\mathbf{\mu}$ & Linear & $Z\times1\times1$ \\
    (12) & (10) & $\mathbf{\sigma}$ & Linear & $Z\times1\times1$ \\
    \midrule
    (13) & (11), (12) & ${\bf z}$ &
    Sample from $\mathcal{N}(\mathbf{\mu}, \mathbf{\sigma})$
    & $Z\times8\times14$ \\
    (14) & (13) & &
    $3\times 3$, stride 1
    & $512\times8\times14$ \\
    (15) & (14) & &
    $\begin{bmatrix}
    3 \times 3, \text{stride 1} \\
    3 \times 3, \text{stride 1}
    \end{bmatrix}$
    & $512\times8\times14$ \\
    (16) & (15) & &
    Upsample $2\times$
    & $512\times16\times28$ \\
    (17) & (16) & &
    $\begin{bmatrix}
    3 \times 3, \text{stride 1} \\
    3 \times 3, \text{stride 1}
    \end{bmatrix}$
    & $256\times16\times28$ \\
    (18) & (17) & &
    Upsample $2\times$
    & $256\times32\times56$ \\
    (19) & (18) & &
    $\begin{bmatrix}
    3 \times 3, \text{stride 1} \\
    3 \times 3, \text{stride 1}
    \end{bmatrix}$
    & $128\times32\times56$ \\
    (20) & (19) & &
    Upsample $2\times$
    & $128\times64\times112$ \\
    (21) & (20) & decoder feat. &
    $\begin{bmatrix}
    3 \times 3, \text{stride 1} \\
    3 \times 3, \text{stride 1}
    \end{bmatrix}$
    & $64\times64\times112$ \\
    \midrule
    (22) & (21) & goal heatmap pred. & $1\times1$, stride 1 & $1\times64\times112$ \\
    \bottomrule
  \end{tabular}
  \vspace{2mm}
  \caption{\small
    \textbf{Overall architecture for our GoalNet.}
    Each convolutional block denoted in the bracket has an internal skip connection with appropriate strides.
    Each convolutional operator is followed by a batch normalization and ReLU layer, except the one before heatmap prediction.
    We denote $N$ as input time frames, $J$ as the number of human keypoints, $Z$ as the dimension of latent space.
    We set $Z = 30$ in our experiments.
    We use nearest upsampling operator.
  }
  \label{tab:supp-goalnet}
  \vspace{-2mm}
\end{table}

\begin{table}
  \centering
  \begin{tabular}{ccccc}
    \toprule
    Index & Input & Data & Operator & Output shape \\
    \midrule
    (1) & - & scene image & - & $3\times256\times448$ \\
    (2) & - & stacked heatmaps & - & $(N \times J)\times256\times448$ \\
    (3) & - & goal heatmap & - & $1\times256\times448$ \\
    (4) & - & initial depth. & - & $N\times1\times1$ \\
    (5) & - & 2D pose sequence & - & $N \times J \times 2$ \\
    \midrule
    (6) & (1), (2), (3) & & $7\times7$, stride 2 & $128\times128\times224$ \\
    (7) & (6) & &
    $\begin{bmatrix}
    3 \times 3, \text{stride 2} \\
    3 \times 3, \text{stride 1}
    \end{bmatrix}$
    & $256\times64\times112$ \\
    (8) & (7) & backbone feat$_1$. & HourglassStack & $256\times64\times112$ \\
    (9) & (8) & backbone feat$_2$. & HourglassStack & $256\times64\times112$ \\
    (10) & (9) & backbone feat$_3$. & HourglassStack & $256\times64\times112$ \\
    \midrule
    (11) & (8) or (9) or (10) & &
    $\begin{bmatrix}
    3 \times 3, \text{stride 1} \\
    3 \times 3, \text{stride 1}
    \end{bmatrix}$
    & $256\times64\times112$ \\
    (12) & (11) & heatmap pred. & $1\times1$, stride 1 & $T\times64\times112$ \\
    \midrule
    (13) & (8) or (9) or (10) & &
    $\begin{bmatrix}
    3 \times 3, \text{stride 2} \\
    3 \times 3, \text{stride 1}
    \end{bmatrix}$
    & $384\times32\times56$ \\
    (14) & (13) & &
    $\begin{bmatrix}
    3 \times 3, \text{stride 2} \\
    3 \times 3, \text{stride 1}
    \end{bmatrix}$
    & $512\times16\times28$ \\
    (15) & (14) & & GlobalAvgPool & $512\times1\times1$ \\
    (16) & (4), (5) & & Linear & $256\times1\times1$ \\
    (17) & (15), (16) & & Linear & $256\times1\times1$ \\
    (18) & (17) & depth pred. & Linear & $(N+T)\times1\times1$ \\
    \bottomrule
  \end{tabular}
  \vspace{2mm}
  \caption{\small
    \textbf{Overall architecture for our PathNet.}
    Each convolutional block denoted in the bracket has an internal skip connection with appropriate strides.
    Each convolutional operator is followed by a batch normalization and ReLU layer, except the one before heatmap prediction.
    Each linear operator is followed by a layer normalization and ReLU layer, except the one before depth prediction.
    We denote $N$ as input time frames, $T$ as output time frames, $J$ as the number of human keypoints.
    We obtain initial depth as input by scaling the size of the human bounding box~\cite{moon2019camera}.
    We define HourglassStack in Table~\ref{tab:supp-hgstack}.
    After each stack output, we use two separate branches for predicting heatmaps and human center depth.
    During training, we backpropagate gradient through all stacks, while during inference, we only use the predictions from final stack.
  }
  \label{tab:supp-pathnet}
  \vspace{-2mm}
\end{table}

\begin{table}
  \centering
  \begin{tabular}{ccccc}
    \toprule
    Index & Input & Data & Operator & Output shape \\
    \midrule
    (1) & - & feat. & - & $256\times64\times112$ \\
    \midrule
    (2) & (1) & &
    $\begin{bmatrix}
    3 \times 3, \text{stride 1} \\
    3 \times 3, \text{stride 1}
    \end{bmatrix}$
    & $256\times64\times112$ \\
    (3) & (2) & &
    $\begin{bmatrix}
    3 \times 3, \text{stride 2} \\
    3 \times 3, \text{stride 1}
    \end{bmatrix}$
    & $384\times32\times56$ \\
    (4) & (3) & &
    $\begin{bmatrix}
    3 \times 3, \text{stride 1} \\
    3 \times 3, \text{stride 1}
    \end{bmatrix}$
    & $384\times32\times56$ \\
    (5) & (4) & &
    $\begin{bmatrix}
    3 \times 3, \text{stride 2} \\
    3 \times 3, \text{stride 1}
    \end{bmatrix}$
    & $512\times16\times28$ \\
    (6) & (5) & &
    $\begin{bmatrix}
    3 \times 3, \text{stride 2} \\
    3 \times 3, \text{stride 1}
    \end{bmatrix}$
    & $512\times8\times14$ \\
    (7) & (6) & &
    $\begin{bmatrix}
    3 \times 3, \text{stride 1} \\
    3 \times 3, \text{stride 1}
    \end{bmatrix}$
    & $512\times8\times14$ \\
    (8) & (7) & &
    $\begin{bmatrix}
    3 \times 3, \text{stride 1} \\
    3 \times 3, \text{stride 1}
    \end{bmatrix}$
    & $512\times8\times14$ \\
    (9) & (8) & &
    Upsample $2\times$
    & $512\times16\times28$ \\
    (10) & (7), (9) & &
    Sum
    & $512\times16\times28$ \\
    (11) & (10) & &
    $\begin{bmatrix}
    3 \times 3, \text{stride 1} \\
    3 \times 3, \text{stride 1}
    \end{bmatrix}$
    & $384\times16\times28$ \\
    (12) & (10) & &
    Upsample $2\times$
    & $384\times32\times56$ \\
    (13) & (4), (12) & &
    Sum
    & $384\times32\times56$ \\
    (14) & (13) & &
    $\begin{bmatrix}
    3 \times 3, \text{stride 1} \\
    3 \times 3, \text{stride 1}
    \end{bmatrix}$
    & $256\times32\times56$ \\
    (15) & (14) & &
    Upsample $2\times$
    & $256\times64\times112$ \\
    \midrule
    (16) & (2), (15) & refined feat. &
    Sum
    & $256\times64\times112$ \\
    \bottomrule
  \end{tabular}
  \vspace{2mm}
  \caption{\small
    \textbf{Modular architecture for one HourglassStack.}
    We follow the design and implementation of \cite{law2019cornernet}.
    Each convolutional block denoted in the bracket has an internal skip connection with appropriate strides.
    We use nearest upsampling operator.
  }
  \label{tab:supp-hgstack}
  \vspace{-2mm}
\end{table}